\pdfoutput=1

\documentclass[11pt]{article}

\usepackage[preprint]{acl}

\usepackage{times}
\usepackage{latexsym}
\usepackage{booktabs}
\usepackage{soul}
\usepackage{amsfonts}
\usepackage{multirow}
\usepackage{rotating}
\usepackage[T1]{fontenc}

\usepackage[utf8]{inputenc}

\usepackage{microtype}

\usepackage{inconsolata}

\usepackage{graphicx}
\usepackage{amsmath}
\usepackage{tabularx}
\usepackage{svg}
\usepackage{subfigure}
\usepackage{graphicx}
\usepackage{float}
\usepackage{hyperref}

\usepackage[nameinlink]{cleveref}

\title{Stands to Reason: Investigating the Effect of Reasoning on Idiomaticity Detection}

\author{
Dylan Phelps\textsuperscript{1, 2},
Rodrigo Wilkens\textsuperscript{3},
Edward Gow-Smith\textsuperscript{1},\\
\textbf{Thomas Pickard}\textsuperscript{1},
\textbf{Maggie Mi}\textsuperscript{1},\and
\textbf{Aline Villavicencio\textsuperscript{1,3}}
\\[0.3cm]
\textsuperscript{1} School of Computer Science, University of Sheffield, UK \\
\textsuperscript{2} Healthy Lifespan Institute, University of Sheffield \\
\textsuperscript{3} Institute of Data Science and Artificial Intelligence, University of Exeter, UK \\
\texttt{\small \{drsphelps1, zmi1, tmrpickard, egow-smith1\}@sheffield.ac.uk} \\
\texttt{\small \{r.wilkens, a.villavicencio\}@exeter.ac.uk,}
}

\begin{document}
\maketitle
\begin{abstract}
The recent trend towards utilisation of reasoning models has improved the performance of Large Language Models (LLMs) across many tasks which involve logical steps. 
One linguistic task that could benefit from this framing is idiomaticity detection, as a potentially idiomatic expression must first be understood before it can be disambiguated and serves as a basis for reasoning. In this paper, we explore how reasoning capabilities in LLMs affect idiomaticity detection performance and examine the effect of model size. We evaluate, as open source representative models, the suite of DeepSeek-R1 distillation models ranging from 1.5B to 70B parameters across four idiomaticity detection datasets. We find the effect of reasoning to be smaller and more varied than expected. For smaller models, producing chain-of-thought (CoT) reasoning increases performance from Math-tuned intermediate models, but not to the levels of the base models, whereas larger models (14B, 32B, and 70B) show modest improvements. Our in-depth analyses reveal that larger models demonstrate good understanding of idiomaticity, successfully producing accurate definitions of expressions, while smaller models often fail to output the actual meaning. For this reason, we also experiment with providing definitions in the prompts of smaller models, which we show can improve performance in some cases.
\end{abstract}

\label{sec:reason}

\section{Introduction}

Large Language Models (LLMs) have been shown to excel at many tasks across many disciplines \citep{ouyang2022training,achiam2023gpt,grattafiori2024llama}, including on tasks involving idiomatic expressions, such as idiomaticity detection and multiword expression identification \citep{phelps_sign_2024, smadu_investigating_2024}. More recently, reasoning models, LLMs which generate a chain of thought (CoT) `reasoning' responses before reaching a final answer \citep{wei_chain--thought_2022}, have been shown to outperform traditional LLMs in a range of tasks \citep{openai_o1_2024}. Notably, DeepSeek-R1 and smaller `distilled' models trained on data produced by it \citep{deepseek-ai_deepseek-r1_2025}, have recently matched and exceeded other reasoning models such as OpenAI's GPT-o1 whilst being open source and offered at a lower price.

The motivation for such reasoning models is that training them to generate outputs in chain-of-thought format allows them ``think'' step-by-step, working out a final answer incrementally. This output format also allows for potentially higher explainability, since the CoT becomes part of the context used to make the final classification. However, some work has shown that CoT explanations can be unfaithful, misrepresenting the reason for a model's prediction \citep{NEURIPS2023_ed3fea90,lyu2023faithful}.

In the field of computational idiom processing, reasoning models are of interest as one could imagine such step-by-step thinking to improve the handling of potentially idiomatic expressions (PIEs), where the meaning is ambiguous and determined by context. Using knowledge of a definition as an indicator of a level of understanding, the CoT also allows analysis of A) how well models can define PIEs in their literal or idiomatic senses, and B) how well models can use subsequent reasoning to work out whether a PIE is used idiomatically or literally in a given context. 

Therefore, in this paper, we explore how the distilled DeepSeek-R1 models perform on a range of idiomaticity detection tasks. Additionally, we evaluate the reasoning outputs of each model to further explore how they represent and understand idiomaticity. Our research questions are: 
\begin{enumerate}
    \item Does the production of reasoning chains improve the ability of models to detect idiomaticity, and how does this vary across model scale?
    \item Do the reasoning outputs (CoT) reflect understanding of the target idiomatic expressions?
    \item Can the definitions produced by the larger models be used as an additional knowledge source, improving the performance of the smaller models?
\end{enumerate}

To this end, we find that although the performance of the models improves as the models scale, the effect of adding reasoning generation is smaller and more varied. Particularly, for the smaller models we find that producing CoTs on average reduces the performance across our datasets, whereas for the larger models a small increase in performance can be seen in the reasoning variants. Our manual analysis of the dataset shows that the larger models also have a good understanding of idiomaticity and can reliably produce accurate definitions of the given expressions, while the smaller models often fail to do so. However, even the largest models are hindered by their ability to use context and reasoning to disambiguate PIEs. 

Our experiments using definitions from the larger models as a knowledge source show that performance of the smallest models improves by an average of 0.069 macro F1 on FLUTE, but does not affect performance significantly for DICE. These results suggest that this methodology has potential as a knowledge distillation technique for certain tasks.

The paper is structured as follows: $\S$\ref{sec-methods} presents the methodology and $\S$\ref{sec:models} presents the results. $\S$\ref{sec:errors} and $\S$\ref{sec:analyses} analyse the results, with $\S$\ref{sec:distil} performing further experiments using the generated definitions.

\section{Methodology}
\label{sec-methods}
In this section, we introduce the datasets, models, and approach that we use in this work.

\subsection{Datasets}

To enable us to compare the performance of the newer models to those evaluated in \citealp{phelps_sign_2024} we evaluate on the same datasets. We also include the recently released DICE dataset \citep{mi_rolling_2024}. We provide brief descriptions of each dataset. 

\paragraph{SemEval 2022 Task 2a \citep{madabushi_semeval-2022_2022}} is a binary classification task for detecting idiomaticity of noun compounds within context sentences, with examples in English, Portuguese, and Galician. The test set contains 150 PIEs split equally across the three languages, with a total of 2342 examples. To maximize the number of test examples, we follow \citealp{phelps_sign_2024} by combining the few-shot and zero-shot test sets. However, when evaluating we don't provide few-shot examples for any of the instances.

\paragraph{Figurative Language Understanding through Textual Explanations (FLUTE; \citealp{chakrabarty_flute_2022})} presents an English-language Natural Language Inference (NLI) task in which models should predict whether a premise, containing a figurative expression, follows from a hypothesis containing a correct or incorrect paraphrase. We evaluate only the idiom subset of the dataset, which contains a test set of 250 examples with 69 idioms represented.

\paragraph{MAGPIE \citep{haagsma_magpie_2020}} is a multi-class idiomaticity detection dataset where a large number of potentially idiomatic expressions in context must be classified as idiomatic, literal, or other with a split of 70/29/1. The entire dataset is in English, with 1134 expressions represented across 4840 examples.

\paragraph{Dataset for Idiomatic Contrastive Evaluation (DICE; \citealp{mi_rolling_2024})} aims to assess the ability of models to use context in idiomaticity detection. Existing datasets fail to assess the role of context in idiom interpretation, as literal meanings often stem from grammatical changes to the form of the idiomatic expression, which allows models to rely on surface cues as a reasoning shortcut instead of true comprehension. To avoid this, DICE contains 2066 sentences that are balanced for sense, in which the form of the expression is kept the same across both literal and figurative uses. Being a newer dataset, the test labels were not publically available when the DeepSeek models were trained, so there is no chance of contamination.

\subsection{Models and Experiments}

\begin{table*}[t!] 
\centering
\begin{tabular}{llrrrr}
\toprule
& Model & Flute & SemEval & MAGPIE & DICE \\
\midrule
\multirow{5}{*}{\begin{turn}{90}Base\end{turn}}  & Qwen2.5-1.5B & 0.849 & 0.458 & 0.430 & 0.366 \\
&Qwen2.5-7B & 0.921 & \textbf{0.737} & 0.786 & 0.710 \\
&Qwen2.5-14B & 0.924 & 0.586 & 0.823 & 0.800 \\
&Qwen2.5-32B & 0.914 & 0.612 & 0.888 & \textbf{0.873} \\
&Llama-70B & 0.921 & 0.658 & 0.778 & 0.816 \\
\midrule
\multirow{2}{*}{\begin{turn}{90}Math\end{turn}}  &  Qwen2.5-Math-1.5B & 0.551 & 0.484 & 0.495 & 0.482 \\
&Qwen2.5-Math-7B & 0.691          & 0.404 & 0.482 & 0.507          \\
\midrule
\multirow{5}{*}{\begin{turn}{90}Reasoning\end{turn}}  &  DeepSeek-R1 Qwen-1.5B & 0.577 & 0.533 & 0.516 & 0.499 \\
&DeepSeek-R1 Qwen-7B   & 0.812          & 0.585 &    0.626  & 0.462          \\
&DeepSeek-R1 Qwen-14B  & 0.929          & 0.573 &  0.863  & 0.863          \\
&DeepSeek-R1 Qwen-32B  & \textbf{0.948} & 0.641 & \textbf{0.890} & \textbf{0.866} \\
&DeepSeek-R1 Llama-70B & \textbf{0.947}          & 0.628 &   0.873  & 0.857          \\
\bottomrule
\end{tabular}
\caption{Results of both the base and the reasoning models on the four datasets. Reported is the mean Macro F1 averaged across 5 runs, using different random seeds. The best score(s) per dataset shown in bold.}
\label{table:main-results}
\end{table*}

The models we evaluate are the suite of DeepSeek-R1 distillation \citep{deepseek-ai_deepseek-r1_2025} models ranging from 1.5-70B parameters, that have been fine-tuned on a reasoning dataset collected from the larger DeepSeek-R1 model (the \textbf{reasoning} models). Whilst processing the input, these models produce chain-of-thought reasoning.

Each of the distilled models is a fine-tuned version of another open source model (the \textbf{base} models), and so to compare the performance with and without CoT reasoning, we also run our evaluations on the base models.
Namely, we evaluate the Qwen2.5 suite of models (1.5B, 7B, 14B, and 32B parameters) \citep{team_qwen25_2024_custom, qwen_qwen25_2025} and their reasoning-tuned DeepSeek-R1 versions, and Llama3.3-70B \citep{llama_llama_2024} alongside its reasoning-tuned variant.

The DeepSeek-R1 1.5B and 7B parameter variants follow a different training pipeline. Rather than direct reasoning-tuning from base Qwen2.5 models. These variants first undergo intermediate training on approximately 1 trillion tokens of mathematical data, with prompting designed to elicit reasoning behaviors \cite{}. This produces three model variants for each size: the original base model (Qwen2.5), the math-specialized intermediate model (Qwen2.5-Math), and the final reasoning-tuned model (DeepSeek-R1 Qwen). We evaluate all three variants to understand the training progression's impact on capabilities. While we anticipate that math-specialization may compromise general domain performance, we include these intermediate models in our evaluation to examine whether subsequent reasoning-tuning with broader domain data can recover the lost general capabilities.

All models are run using the vLLM library \citep{kwon2023efficient} utilizing Q6\_K\_M quantizations of each model \citep{frantar_gptq_2023}. This allows all the models to be run on a single A100 80GB GPU.

For each dataset, we prompt the models with the prompts given in \Cref{sec:prompts}. The reasoning models generate a CoT before the output, which we split off for analysis. Where the models do not output the required label in an easily parseable format, we use GPT-4o \citep{openai2024gpt4ocard} to extract the returned label (using the prompt also given in \Cref{sec:prompts}). We evaluate the models using Macro F1 for all datasets.

\section{Model Performance}
\label{sec:models}

We present the results from our experiments in \Cref{table:main-results}. As expected, we see the larger models achieving the highest performance on the idiomaticity detection tasks. The best performing model is DeepSeek-R1 Qwen-32B, which slightly outperforms the larger DeepSeek-R1 Llama-70B  on all of the datasets. The only model that surpasses comparable larger models is the base version of Qwen2.5-7b, which outperforms the larger non-reasoning models on FLUTE, and, surprisingly, gets the highest score on SemEval-2022, being the only model to score over 0.7.

In relation to the first question, if  reasoning improves the ability of models to detect idiomaticity, the results observed paint a mostly positive picture. In \Cref{tab:model-diff} we show the difference between the base and reasoning variants of each model across the datasets, as well as the difference between the base and intermediate math models: Qwen2.5-Math-1b and Qwen2.5-Math-7b.

\begin{table*}
\centering
\begin{tabular}{llrrrrrrrr}
\toprule
& \multicolumn{1}{l}{Model} & \multicolumn{1}{l}{FLUTE} & \multicolumn{4}{c}{SemEval} & \multicolumn{1}{l}{MAGPIE} & \multicolumn{1}{l}{DICE} & \multicolumn{1}{l}{\textbf{Mean Diff.}} \\
 \cmidrule(lr){4-7}
 & & \multicolumn{1}{l}{} & \multicolumn{1}{l}{ALL} & \multicolumn{1}{l}{EN} & \multicolumn{1}{l}{PT} & \multicolumn{1}{l}{GL} & \multicolumn{1}{l}{} & \multicolumn{1}{l}{} & \multicolumn{1}{l}{} \\ 
 \midrule
\multirow{5}{*}{\begin{turn}{90}Base\end{turn}} & 1.5b & -0.272 & 0.075 & 0.075 & 0.068 & 0.082 & 0.086 & 0.132 & 0.006 \\
& 7b & -0.109 & -0.153 & -0.107 & -0.138 & -0.077 & -0.160 & -0.247 & -0.167 \\
& 14b & 0.005 & -0.014 & -0.015 & 0.051 & 0.011 & 0.040 & 0.064 & 0.024 \\
& 32b & 0.034 & 0.039 & 0.039 & 0.081 & -0.010 & 0.002 & -0.007 & 0.017 \\
& 70b & 0.026 & -0.021 & -0.066 & -0.085 & -0.019 & 0.095 & 0.042 & 0.035 \\
\midrule
\multirow{2}{*}{\begin{turn}{90}Math\end{turn}} &1.5b & 0.026 & 0.049 & 0.050 & 0.050 & 0.057 & 0.021 & 0.017 & 0.028 \\
& 7b & 0.121 & 0.181 & 0.212 & 0.231 & 0.206 & 0.115 & -0.044 & 0.093 \\

\bottomrule
\end{tabular}

\caption{Difference in performance of the reasoning models compared to `Base' and `Math' variants, in absolute difference in Macro F1. A negative value means reasoning hurts performance, whilst a positive value means it improves performance. The ``Mean Diff.'' column reports the average improvement from using a reasoning model across the four datasets.}
\label{tab:model-diff}
\end{table*}

Math-tuning produces the large drops in performance we expected for both the Qwen2.5-1b and Qwen2.5-7b models. This is slightly less pronounced on the 1.5b model, though this primarily reflects floor effects as performance on SemEval, MAGPIE, and DICE was already near-random levels, leaving little room for further deterioration. Conversely, the reasoning tuning has a large positive effect when moving from the Math variants to the DeepSeek variants, with overall positive improvements of 0.028 and 0.093, an effect that can especially be seen through single task improvements for the 7b parameter model (0.121, 0.181, 0.115). 

Larger models (14B, 32B, and 70B) demonstrate consistent benefits from CoT reasoning integration, with performance improvements reaching 0.095 for Llama-70B on MAGPIE. However, results vary in magnitude across tasks, and some performance decreases occur on the SemEval dataset—particularly for Portuguese and Galician languages, where the effect of reasoning is less clear.

\section{Manual Error Analysis}
\label{sec:errors}

As an initial step in exploring the understanding of the models, we manually inspect samples of the reasoning outputs and evaluate them. We split the evaluation into two facets: whether the model understands the relevant PIE (whether it generates a valid idiomatic definition, either wholly or in parts throughout the CoT), and whether the reasoning is valid (whether the model successfully disambiguates the PIE in context within its CoT). This categorization allows us to separate the effects of the model's understanding of the expression and context and its reasoning ability. 

\subsection{Labelling Setup}

\begin{figure*}[ht!]
    \centering
    \includegraphics[width=\textwidth]{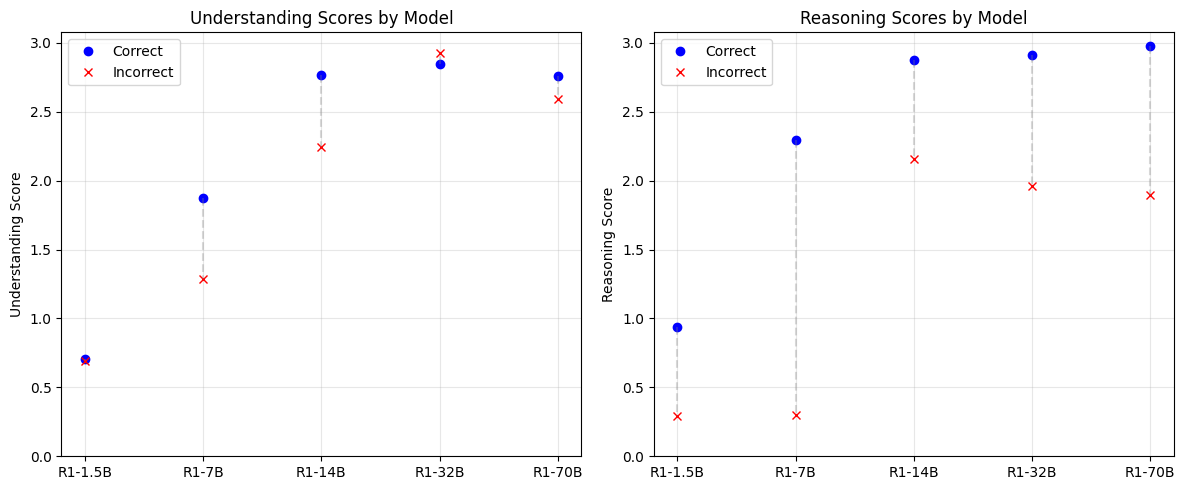}
  \caption{The average manually labelled reasoning and understanding scores (between 0 and 3) for each model, for 15 correct and 15 incorrect examples, averaged across DICE and MAGPIE.}
  \label{fig:labelling-analysis}
\end{figure*}

Three of the authors, acting as annotators, each annotate the same 30 responses (15 correct and 15 incorrect) for each reasoning model. We initially chose responses to the MAGPIE dataset as this is the largest dataset, with the largest variance in expressions used. However, whilst labelling we found the dataset to be quite noisy (mislabelling, example sentences which do not actually contain the target expression, etc.) and so we additionally label examples from DICE. Overall, 300 examples were labelled 3 times (once by each annotator) for understanding and reasoning ability on a 5-point scale. Each point on the scale was awarded a score, which is averaged across annotators and examples, to allow us to get a representative score for the model. The full set of labels and scores can be seen in \Cref{sec:error-analysis-labels}.

For understanding, labels range from No definition, Focus on one word, and Inaccurate definition (all scored 0 for lacking accurate information) to Partial definition (scored 2) and Accurate definition (scored 3). Reasoning uses a similar scale: Nonsensical reasoning (0), OK reasoning that partially attends to expression and context (2), and Good reasoning with mostly correct connections (3). We label meta-reasoning, which occurs (rarely) when the model doesn't correctly identify the expression in the context, instead relying on meta-knowledge typical expression usage. We use 0-2-3 scoring instead of 1-5 to emphasize the gap between inaccurate and accurate responses, with some labels being assigned the same score as they identify different scenarios of similar quality.

\subsection{Labelling Findings}

\subsubsection{Mislabelling in the dataset}

The first thing we observe whilst manually labelling is that, of the examples where the model is judged to be incorrect, there are many labels which are either incorrect or ambiguous. In \Cref{tab:mislabel}, we show the proportion of examples where the gold label and prediction do not match, for which our annotators agreed that the gold label was either incorrect or ambiguous. There is some overlap between the examples, as he idiomatic subset of DICE (which accounts for most of these examples) is taken from MAGPIE. For the larger models, this effect is particularly pronounced, with a large proportion of their ``incorrect'' predictions actually being judged valid by human annotators. This indicates that accuracy for the larger models may be higher than suggested by the results in \Cref{table:main-results}, which further reinforces the high level of understanding shown by the larger models. We emphasise that this does not imply a large proportion of these datasets are incorrectly labelled or ambiguous, instead the accuracy of the large models means that when they are incorrect they have a high chance of picking out the few examples that are ambiguous.

\begin{table}[t!]
\centering
\begin{tabular}{lrr}
\toprule
 & \multicolumn{2}{c}{\% Judged Valid}  \\ \cmidrule{2-3} 
      & \multicolumn{1}{c}{DICE} & MAGPIE \\ \midrule
DeepSeek-R1 Qwen-1.5B & \multicolumn{1}{r}{0\%}    & 0\%      \\
DeepSeek-R1 Qwen-7B   & \multicolumn{1}{r}{0\%}    & 6.7\%    \\
DeepSeek-R1 Qwen-14B  & \multicolumn{1}{r}{40\%}   & 60\%     \\
DeepSeek-R1 Qwen-32B  & \multicolumn{1}{r}{46.7\%} & 33.3\%   \\
DeepSeek-R1 Llama-70B & \multicolumn{1}{r}{40\%}   & 33.3\%   \\
\bottomrule
\end{tabular}
\caption{For each model, for DICE and MAPGIE, the percentage of sampled ``incorrect'' predictions (where the model prediction does not match the gold label), where the model prediction was judged to be valid by the annotators, due to a mislabelled or ambiguous example.}
\label{tab:mislabel}
\end{table}

\subsubsection{Error Patterns by Model Size}

\begin{table*}
\centering
\setlength{\extrarowheight}{0pt}
\addtolength{\extrarowheight}{\aboverulesep}
\addtolength{\extrarowheight}{\belowrulesep}
\setlength{\aboverulesep}{0pt}
\setlength{\belowrulesep}{0pt}

\begin{tabular}{lrrrrrr} 
\toprule
Model         & \multicolumn{1}{l}{Flute}          & \multicolumn{3}{c}{SemEval}                                                                                  & \multicolumn{1}{l}{MAGPIE}         & \multicolumn{1}{l}{DICE}            \\ 
\cmidrule(l){3-5}
              & \multicolumn{1}{l}{}               & \multicolumn{1}{l}{EN}             & \multicolumn{1}{l}{PT}             & \multicolumn{1}{l}{GL}             & \multicolumn{1}{l}{}               & \multicolumn{1}{l}{}                \\ 
\midrule
DeepSeek 1.5B & -0.038 & 0.009  & -0.037 & -0.061 & 0.000      & -0.011  \\
DeepSeek 7B   & \textbf{-0.151} & \textbf{-0.075} & -0.001 & \textbf{-0.089} & \textbf{-0.057} & -0.006  \\
DeepSeek 14B  & -0.022 & 0.020   & -0.036 & 0.000      & \textbf{-0.135} & \textbf{-0.177}  \\
DeepSeek 32B  & \textbf{-0.246} & -0.003 & -0.064 & \textbf{0.126}  & \textbf{-0.119} & \textbf{-0.147}  \\
DeepSeek 70B  & \textbf{-0.305} & -0.036 & -0.002 & 0.028  & \textbf{-0.132} & \textbf{-0.156}  \\

\bottomrule
\end{tabular}
\caption{Correlation between CoT length and model correctness. Values in bold indicate a p-value $<$ 0.05.}
\label{tab:corr_model_CoTlen}
\end{table*}

\Cref{fig:labelling-analysis} shows the manual label scores for each model in the case where the model is correct and incorrect. We observe that mistakes made by different model sizes often fall into different categories of errors. The larger models appear to have an understanding of most of the expressions in the data, and therefore as part of their reasoning produce a definition of the idiom. The smaller models, however, often produce an incorrect definition or struggle to identify or recognize the expression all together. This then leaves a poor basis for the rest of the reasoning, leaving the label as almost random.

The 1.5B parameter model shows a low level of both understanding and reasoning ability, with most of its correct outputs being due to chance. The 7B parameter model has on average higher levels of both reasoning and understanding, which correlates with the results in \Cref{table:main-results}. The main factor to whether the model is correct or incorrect appears to be the reasoning performance, as this varies from close to the larger models when correct, but closer to the 1.5B model when incorrect. The larger models understanding of the expressions is generally high when it is both correct and incorrect, with more variation being identified in the reasoning capabilities. However, the overall reasoning scores are still relatively high, which may indicate that they are incorrect on the more nuanced examples, something we noted whilst labelling.





\section{Chain-of-Thought Length}
\label{sec:analyses}

\begin{figure}
    \centering
    \includegraphics[width=0.7\linewidth]{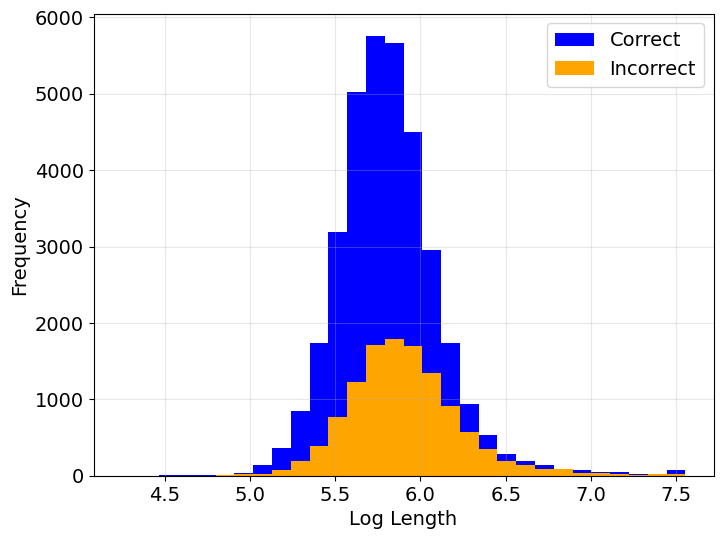}
    \caption{The distribution of reasoning output log lengths for correct (blue) vs. incorrect (orange) model predictions, aggregated across all models and datasets.}
    \label{fig:hist_cot_len}
\end{figure}

We investigate the potential relationship between the Chain of Thought (CoT) size and model accuracy. The DeepSeek paper showed that as model training progressed and model performance improved, average CoT length went up \citep{deepseek-ai_deepseek-r1_2025}.
To explore the correlation between reasoning length and performance in idiomaticity detection, we split the CoTs into two categories for incorrect and correct predictions, and then tokenise them (using spaCy + en\_core\_web\_sm) to give lengths. We show two histograms of CoT length for correct and incorrect predictions in \Cref{fig:hist_cot_len}. We can see that the CoT length distribution appears similar across both cases. Next, we calculate the point-biserial correlation between CoT size and prediction correctness for each model size category (1.5B, 7B, 14B, 32B and 70B), which can be seen in Table \ref{tab:corr_model_CoTlen}. Additionally, we compute the pseudo-R² for a logistic regression model that predicts accuracy based on CoT length.

The results reveal that, across various datasets, there is no consistent pattern suggesting a significant relationship between CoT size and prediction accuracy. For instance, the most considerable observed correlation (-30.50\%) in the EN-FLUTE corpus corresponds to a modest R\textsuperscript{2} value of 0.11, meaning only 11\% of the variance in prediction outcomes is explained by CoT size. While models of larger sizes (32B and 70B) show slight improvements in R² for certain corpora, such as FLUTE and DICE, the overall findings suggest no clear trend linking CoT length to predictive success.

\section{Definition Generation as Distillation}
\label{sec:distil}

\begin{table*}[h!]
\centering
\begin{tabular}{lcccc}
\toprule
\multicolumn{1}{l}{Model} & \multicolumn{2}{c}{Original Prompt}  & \multicolumn{2}{c}{Definition Prompt} \\
\cmidrule(lr){2-3}  \cmidrule(lr){4-5}
& \multicolumn{1}{c}{FLUTE} & \multicolumn{1}{c}{DICE} & \multicolumn{1}{c}{FLUTE} & \multicolumn{1}{c}{DICE}  \\ 
\midrule
DeepSeek-R1 Qwen-1.5B & 0.577 & 0.499 & \textbf{0.663} & 0.487 \\
DeepSeek-R1 Qwen-7B & 0.812 & 0.462 & \textbf{0.864} & 0.461 \\
DeepSeek-R1 Qwen-14B & 0.929 & 0.863 & 0.908 & 0.842 \\
\bottomrule
\end{tabular}
\caption{The results using both the original and definition appended prompt, when evaluating the smaller models on FLUTE and DICE. We report the mean Macro F1 across 5 runs with different random seeds, and bold any significantly improved results for each model/dataset pair.}
\label{tab:def-results}
\end{table*}

\begin{table}\centering
\resizebox{\columnwidth}{!}{
\begin{tabular}{lp{0.7\linewidth}}
\toprule
PIE & Generated Definition \\
\midrule
all hell broke loose & ``All hell broke loose'' means that a situation suddenly became extremely chaotic, noisy, or violent. \\
off the hook & To be off the hook means to be free from a responsibility, obligation, or troublesome situation, often after being expected to handle it.  \\
against the grain & To go against the grain is to act in a way that is contrary to the usual approach, often resulting in resistance.  \\
play with fire & `Play with fire' means to engage in a risky or dangerous situation that could lead to negative consequences. \\
make a killing & To make a killing means to achieve great success or earn a lot of money. \\
\bottomrule
\end{tabular}}
\caption{Examples of the definitions generated by DeepSeek-R1 32B for given PIEs from the DICE dataset.}
\label{tab:defs}
\end{table}

The results from our manual labelling reveal that the 32B and 70B parameter models can accurately produce definitions of the expressions represented in the dataset. Given the lack of comprehensive resources for idiomatic expression definitions, this finding presents an opportunity to use model-generated definitions as a novel distillation technique.

As an initial exploration, we investigate whether providing the model generated definitions from the larger models to the smaller reasoning models can improve performance. In our manual analysis, we found that the 1.5B and 7B parameter reasoning models were not able to accurately reproduce definitions, which may suggest that they do not contain the required knowledge to do so, partly explaining their poor performance on the detection tasks. By providing this information in the prompt, this knowledge is able to be used by the models whilst making predictions.

For this experiment, we use DeepSeek-R1 32B to generate definitions for all the expressions in the FLUTE and DICE datasets (examples shown in Table \ref{tab:defs}). We select DeepSeek-R1 32B as this shows the highest overall `understanding' score in the manual analysis (see Figure \ref{fig:labelling-analysis}). We then append the definition to the prompt used in our main experiments and rerun the evaluation for the 1.5B, 7B, and 14B parameter reasoning models. These models represent a range of `understanding' scores, allowing us to analyse the effectiveness of the definitions across different performance and `understanding' levels. We expect that the models with the lowest performance on the detection tasks and `understanding', e.g. the 1.5B and 7B parameter models, will benefit the most from the additional information in the prompt.

\subsection{Results}

The results from the experiment (shown in Table \ref{tab:def-results}) show that adding definitions to the prompt significantly increases performance for the 1.5B and 7B parameter models on the FLUTE dataset. However, for DICE, and for the 14B parameter model on both datasets, no significant difference can be seen between adding the definition to the prompt or not.

We suggest that the difference in effectiveness on FLUTE and DICE comes from the fact that, as suggested by \cite{mi_rolling_2024}, DICE requires models to attend to the surrounding context to make decisions. Adding the definition will not improve the models' ability to do this, and so no increase in performance is seen.

However, the lack of any decrease in performance is promising for the technique as it means that adding definitions to the prompt can safely be done without risk of reducing performance. While the 14B parameter model shows slightly larger absolute decreases on both datasets, these differences are not statistically significant.

\section{Conclusion}
In this study, we have investigated the effect that reasoning has on the idiomaticity detection ability of LLMs. On four idiomaticity datasets (Flute, SemEval, MAGPIE, DICE), we have evaluated the performance of the distilled DeepSeek-R1 models across a range of sizes, in addition to their corresponding non-reasoning base models, and intermediate Math variants for the small parameter count models. To further explore the results, we have performed both a manual and a quantitative analysis on the reasoning outputs of the models in order to gain some insight into the models' understanding of idiomaticity, as well as their ability to reason with the expression in context. As a result, we have three main findings: 
\begin{enumerate}
    \item The effect of reasoning on idiomaticity detection is relatively small, although it varies across model size and per dataset. For the smaller models, training on math specific reasoning data greatly reduces performance, but subsequent training on more general domain reasoning data can restore performance. Larger models show slight increases in performance when trained on general domain reasoning data.
    \item Our manual analysis shows the larger models are consistently able to produce high quality definitions of the expressions, something which the smaller models struggle to do. However, for all models the quality of the reasoning is the most prominent factor which decides whether a model will be correct or not. This highlights that even the largest models struggle to use the expression and the context to disambiguate the meaning.
    \item Our further experiments utilising the definitions produced by the larger models as knowledge sources for smaller models show promise as a knowledge distillation technique. Appending the definitions to the prompt improves the performance in some cases, whilst having no significant effect in others, implying this technique can be applied generally without risk of regression.
\end{enumerate}

Our results show that while CoT reasoning offers small benefits for idiomaticity detection, the modest improvements suggest that current reasoning approaches may not fully address the underlying contextual disambiguation challenges inherent in idiomaticity detection. Even if the reasoning generated appears to indicate idiomatic understanding, further examination indicates otherwise. Future work should explore alternative reasoning frameworks specifically designed for figurative language processing, perhaps incorporating more structured approaches to context interpretation. Additionally, it could explore other ways to utilise the high-quality definitions generated by the models, either through in-context learning or as fine-tuning data.

\bibliography{custom, references}

\begin{thebibliography}{20}
\providecommand{\natexlab}[1]{#1}

\bibitem[{Achiam et~al.(2023)Achiam, Adler, Agarwal, Ahmad, Akkaya, Aleman, Almeida, Altenschmidt, Altman, Anadkat et~al.}]{achiam2023gpt}
Josh Achiam, Steven Adler, Sandhini Agarwal, Lama Ahmad, Ilge Akkaya, Florencia~Leoni Aleman, Diogo Almeida, Janko Altenschmidt, Sam Altman, Shyamal Anadkat, et~al. 2023.
\newblock Gpt-4 technical report.
\newblock \emph{arXiv preprint arXiv:2303.08774}.

\bibitem[{Chakrabarty et~al.(2022)Chakrabarty, Saakyan, Ghosh, and Muresan}]{chakrabarty_flute_2022}
Tuhin Chakrabarty, Arkadiy Saakyan, Debanjan Ghosh, and Smaranda Muresan. 2022.
\newblock \href {https://doi.org/10.18653/v1/2022.emnlp-main.481} {{FLUTE}: {Figurative} {Language} {Understanding} through {Textual} {Explanations}}.
\newblock In \emph{Proceedings of the 2022 {Conference} on {Empirical} {Methods} in {Natural} {Language} {Processing}}, pages 7139--7159, Abu Dhabi, United Arab Emirates. Association for Computational Linguistics.

\bibitem[{DeepSeek-AI et~al.(2025)DeepSeek-AI, Guo, Yang, Zhang, Song, Zhang, Xu, Zhu, Ma, Wang, Bi, Zhang, Yu, Wu, Wu, Gou, Shao, Li, Gao, Liu, Xue, Wang, Wu, Feng, Lu, Zhao, Deng, Zhang, Ruan, Dai, Chen, Ji, Li, Lin, Dai, Luo, Hao, Chen, Li, Zhang, Bao, Xu, Wang, Ding, Xin, Gao, Qu, Li, Guo, Li, Wang, Chen, Yuan, Qiu, Li, Cai, Ni, Liang, Chen, Dong, Hu, Gao, Guan, Huang, Yu, Wang, Zhang, Zhao, Wang, Zhang, Xu, Xia, Zhang, Zhang, Tang, Li, Wang, Li, Tian, Huang, Zhang, Wang, Chen, Du, Ge, Zhang, Pan, Wang, Chen, Jin, Chen, Lu, Zhou, Chen, Ye, Wang, Yu, Zhou, Pan, Li, Zhou, Wu, Ye, Yun, Pei, Sun, Wang, Zeng, Zhao, Liu, Liang, Gao, Yu, Zhang, Xiao, An, Liu, Wang, Chen, Nie, Cheng, Liu, Xie, Liu, Yang, Li, Su, Lin, Li, Jin, Shen, Chen, Sun, Wang, Song, Zhou, Wang, Shan, Li, Wang, Wei, Zhang, Xu, Li, Zhao, Sun, Wang, Yu, Zhang, Shi, Xiong, He, Piao, Wang, Tan, Ma, Liu, Guo, Ou, Wang, Gong, Zou, He, Xiong, Luo, You, Liu, Zhou, Zhu, Xu, Huang, Li, Zheng, Zhu, Ma, Tang, Zha, Yan, Ren, Ren, Sha, Fu, Xu, Xie, Zhang,
  Hao, Ma, Yan, Wu, Gu, Zhu, Liu, Li, Xie, Song, Pan, Huang, Xu, Zhang, and Zhang}]{deepseek-ai_deepseek-r1_2025}
DeepSeek-AI, Daya Guo, Dejian Yang, Haowei Zhang, Junxiao Song, Ruoyu Zhang, Runxin Xu, Qihao Zhu, Shirong Ma, Peiyi Wang, Xiao Bi, Xiaokang Zhang, Xingkai Yu, Yu~Wu, Z.~F. Wu, Zhibin Gou, Zhihong Shao, Zhuoshu Li, Ziyi Gao, Aixin Liu, Bing Xue, Bingxuan Wang, Bochao Wu, Bei Feng, Chengda Lu, Chenggang Zhao, Chengqi Deng, Chenyu Zhang, Chong Ruan, Damai Dai, Deli Chen, Dongjie Ji, Erhang Li, Fangyun Lin, Fucong Dai, Fuli Luo, Guangbo Hao, Guanting Chen, Guowei Li, H.~Zhang, Han Bao, Hanwei Xu, Haocheng Wang, Honghui Ding, Huajian Xin, Huazuo Gao, Hui Qu, Hui Li, Jianzhong Guo, Jiashi Li, Jiawei Wang, Jingchang Chen, Jingyang Yuan, Junjie Qiu, Junlong Li, J.~L. Cai, Jiaqi Ni, Jian Liang, Jin Chen, Kai Dong, Kai Hu, Kaige Gao, Kang Guan, Kexin Huang, Kuai Yu, Lean Wang, Lecong Zhang, Liang Zhao, Litong Wang, Liyue Zhang, Lei Xu, Leyi Xia, Mingchuan Zhang, Minghua Zhang, Minghui Tang, Meng Li, Miaojun Wang, Mingming Li, Ning Tian, Panpan Huang, Peng Zhang, Qiancheng Wang, Qinyu Chen, Qiushi Du, Ruiqi Ge, Ruisong
  Zhang, Ruizhe Pan, Runji Wang, R.~J. Chen, R.~L. Jin, Ruyi Chen, Shanghao Lu, Shangyan Zhou, Shanhuang Chen, Shengfeng Ye, Shiyu Wang, Shuiping Yu, Shunfeng Zhou, Shuting Pan, S.~S. Li, Shuang Zhou, Shaoqing Wu, Shengfeng Ye, Tao Yun, Tian Pei, Tianyu Sun, T.~Wang, Wangding Zeng, Wanjia Zhao, Wen Liu, Wenfeng Liang, Wenjun Gao, Wenqin Yu, Wentao Zhang, W.~L. Xiao, Wei An, Xiaodong Liu, Xiaohan Wang, Xiaokang Chen, Xiaotao Nie, Xin Cheng, Xin Liu, Xin Xie, Xingchao Liu, Xinyu Yang, Xinyuan Li, Xuecheng Su, Xuheng Lin, X.~Q. Li, Xiangyue Jin, Xiaojin Shen, Xiaosha Chen, Xiaowen Sun, Xiaoxiang Wang, Xinnan Song, Xinyi Zhou, Xianzu Wang, Xinxia Shan, Y.~K. Li, Y.~Q. Wang, Y.~X. Wei, Yang Zhang, Yanhong Xu, Yao Li, Yao Zhao, Yaofeng Sun, Yaohui Wang, Yi~Yu, Yichao Zhang, Yifan Shi, Yiliang Xiong, Ying He, Yishi Piao, Yisong Wang, Yixuan Tan, Yiyang Ma, Yiyuan Liu, Yongqiang Guo, Yuan Ou, Yuduan Wang, Yue Gong, Yuheng Zou, Yujia He, Yunfan Xiong, Yuxiang Luo, Yuxiang You, Yuxuan Liu, Yuyang Zhou, Y.~X. Zhu,
  Yanhong Xu, Yanping Huang, Yaohui Li, Yi~Zheng, Yuchen Zhu, Yunxian Ma, Ying Tang, Yukun Zha, Yuting Yan, Z.~Z. Ren, Zehui Ren, Zhangli Sha, Zhe Fu, Zhean Xu, Zhenda Xie, Zhengyan Zhang, Zhewen Hao, Zhicheng Ma, Zhigang Yan, Zhiyu Wu, Zihui Gu, Zijia Zhu, Zijun Liu, Zilin Li, Ziwei Xie, Ziyang Song, Zizheng Pan, Zhen Huang, Zhipeng Xu, Zhongyu Zhang, and Zhen Zhang. 2025.
\newblock \href {https://doi.org/10.48550/arXiv.2501.12948} {{DeepSeek}-{R1}: {Incentivizing} {Reasoning} {Capability} in {LLMs} via {Reinforcement} {Learning}}.
\newblock \emph{arXiv preprint}.
\newblock ArXiv:2501.12948 [cs].

\bibitem[{Frantar et~al.(2023)Frantar, Ashkboos, Hoefler, and Alistarh}]{frantar_gptq_2023}
Elias Frantar, Saleh Ashkboos, Torsten Hoefler, and Dan Alistarh. 2023.
\newblock \href {https://doi.org/10.48550/arXiv.2210.17323} {{GPTQ}: {Accurate} {Post}-{Training} {Quantization} for {Generative} {Pre}-trained {Transformers}}.
\newblock ICLR 2023.
\newblock ArXiv:2210.17323 [cs].

\bibitem[{Grattafiori et~al.(2024)Grattafiori, Dubey, Jauhri, Pandey, Kadian, Al-Dahle, Letman, Mathur, Schelten, Vaughan et~al.}]{grattafiori2024llama}
Aaron Grattafiori, Abhimanyu Dubey, Abhinav Jauhri, Abhinav Pandey, Abhishek Kadian, Ahmad Al-Dahle, Aiesha Letman, Akhil Mathur, Alan Schelten, Alex Vaughan, et~al. 2024.
\newblock The llama 3 herd of models.
\newblock \emph{arXiv preprint arXiv:2407.21783}.

\bibitem[{Haagsma et~al.(2020)Haagsma, Bos, and Nissim}]{haagsma_magpie_2020}
Hessel Haagsma, Johan Bos, and Malvina Nissim. 2020.
\newblock \href {https://aclanthology.org/2020.lrec-1.35/} {{MAGPIE}: {A} {Large} {Corpus} of {Potentially} {Idiomatic} {Expressions}}.
\newblock In \emph{Proceedings of the {Twelfth} {Language} {Resources} and {Evaluation} {Conference}}, pages 279--287, Marseille, France. European Language Resources Association.

\bibitem[{Kwon et~al.(2023)Kwon, Li, Zhuang, Sheng, Zheng, Yu, Gonzalez, Zhang, and Stoica}]{kwon2023efficient}
Woosuk Kwon, Zhuohan Li, Siyuan Zhuang, Ying Sheng, Lianmin Zheng, Cody~Hao Yu, Joseph~E. Gonzalez, Hao Zhang, and Ion Stoica. 2023.
\newblock Efficient memory management for large language model serving with pagedattention.
\newblock In \emph{Proceedings of the ACM SIGOPS 29th Symposium on Operating Systems Principles}.

\bibitem[{{Llama Team}(2024)}]{llama_llama_2024}
{Llama Team}. 2024.
\newblock \href {https://doi.org/10.48550/arXiv.2407.21783} {The {Llama} 3 {Herd} of {Models}}.
\newblock \emph{arXiv preprint}.
\newblock ArXiv:2407.21783 [cs].

\bibitem[{Lyu et~al.(2023)Lyu, Havaldar, Stein, Zhang, Rao, Wong, Apidianaki, and Callison-Burch}]{lyu2023faithful}
Qing Lyu, Shreya Havaldar, Adam Stein, Li~Zhang, Delip Rao, Eric Wong, Marianna Apidianaki, and Chris Callison-Burch. 2023.
\newblock Faithful chain-of-thought reasoning.
\newblock In \emph{The 13th International Joint Conference on Natural Language Processing and the 3rd Conference of the Asia-Pacific Chapter of the Association for Computational Linguistics (IJCNLP-AACL 2023)}.

\bibitem[{Madabushi et~al.(2022)Madabushi, Gow-Smith, Garcia, Scarton, Idiart, and Villavicencio}]{madabushi_semeval-2022_2022}
Harish~Tayyar Madabushi, Edward Gow-Smith, Marcos Garcia, Carolina Scarton, Marco Idiart, and Aline Villavicencio. 2022.
\newblock \href {https://doi.org/10.48550/arXiv.2204.10050} {{SemEval}-2022 {Task} 2: {Multilingual} {Idiomaticity} {Detection} and {Sentence} {Embedding}}.
\newblock \emph{arXiv preprint}.
\newblock Number: arXiv:2204.10050 arXiv:2204.10050 [cs].

\bibitem[{Mi et~al.(2024)Mi, Villavicencio, and Moosavi}]{mi_rolling_2024}
Maggie Mi, Aline Villavicencio, and Nafise~Sadat Moosavi. 2024.
\newblock \href {https://doi.org/10.48550/ARXIV.2410.16069} {Rolling the {DICE} on {Idiomaticity}: {How} {LLMs} {Fail} to {Grasp} {Context}}.
\newblock Publisher: arXiv Version Number: 1.

\bibitem[{OpenAI(2024)}]{openai2024gpt4ocard}
OpenAI. 2024.
\newblock \href {https://arxiv.org/abs/2410.21276} {Gpt-4o system card}.
\newblock \emph{Preprint}, arXiv:2410.21276.

\bibitem[{OpenAI(2025)}]{openai_o1_2024}
OpenAI. 2025.
\newblock \href {https://openai.com/index/learning-to-reason-with-llms/} {Learning to reason with {LLMs}}.

\bibitem[{Ouyang et~al.(2022)Ouyang, Wu, Jiang, Almeida, Wainwright, Mishkin, Zhang, Agarwal, Slama, Ray et~al.}]{ouyang2022training}
Long Ouyang, Jeffrey Wu, Xu~Jiang, Diogo Almeida, Carroll Wainwright, Pamela Mishkin, Chong Zhang, Sandhini Agarwal, Katarina Slama, Alex Ray, et~al. 2022.
\newblock Training language models to follow instructions with human feedback.
\newblock \emph{Advances in neural information processing systems}, 35:27730--27744.

\bibitem[{Phelps et~al.(2024)Phelps, Pickard, Mi, Gow-Smith, and Villavicencio}]{phelps_sign_2024}
Dylan Phelps, Thomas M.~R. Pickard, Maggie Mi, Edward Gow-Smith, and Aline Villavicencio. 2024.
\newblock \href {https://aclanthology.org/2024.mwe-1.22/} {Sign of the {Times}: {Evaluating} the use of {Large} {Language} {Models} for {Idiomaticity} {Detection}}.
\newblock In \emph{Proceedings of the {Joint} {Workshop} on {Multiword} {Expressions} and {Universal} {Dependencies} ({MWE}-{UD}) @ {LREC}-{COLING} 2024}, pages 178--187, Torino, Italia. ELRA and ICCL.

\bibitem[{Qwen(2024)}]{team_qwen25_2024_custom}
Qwen. 2024.
\newblock \href {https://qwenlm.github.io/blog/qwen2.5/} {Qwen2.5: {A} {Party} of {Foundation} {Models}!}
\newblock Section: blog.

\bibitem[{Qwen et~al.(2025)Qwen, Yang, Yang, Zhang, Hui, Zheng, Yu, Li, Liu, Huang, Wei, Lin, Yang, Tu, Zhang, Yang, Yang, Zhou, Lin, Dang, Lu, Bao, Yang, Yu, Li, Xue, Zhang, Zhu, Men, Lin, Li, Tang, Xia, Ren, Ren, Fan, Su, Zhang, Wan, Liu, Cui, Zhang, and Qiu}]{qwen_qwen25_2025}
Qwen, An~Yang, Baosong Yang, Beichen Zhang, Binyuan Hui, Bo~Zheng, Bowen Yu, Chengyuan Li, Dayiheng Liu, Fei Huang, Haoran Wei, Huan Lin, Jian Yang, Jianhong Tu, Jianwei Zhang, Jianxin Yang, Jiaxi Yang, Jingren Zhou, Junyang Lin, Kai Dang, Keming Lu, Keqin Bao, Kexin Yang, Le~Yu, Mei Li, Mingfeng Xue, Pei Zhang, Qin Zhu, Rui Men, Runji Lin, Tianhao Li, Tianyi Tang, Tingyu Xia, Xingzhang Ren, Xuancheng Ren, Yang Fan, Yang Su, Yichang Zhang, Yu~Wan, Yuqiong Liu, Zeyu Cui, Zhenru Zhang, and Zihan Qiu. 2025.
\newblock \href {https://doi.org/10.48550/arXiv.2412.15115} {Qwen2.5 {Technical} {Report}}.
\newblock \emph{arXiv preprint}.
\newblock ArXiv:2412.15115 [cs].

\bibitem[{Smădu et~al.(2024)Smădu, Ion, Cercel, Pop, and Cercel}]{smadu_investigating_2024}
Răzvan-Alexandru Smădu, David-Gabriel Ion, Dumitru-Clementin Cercel, Florin Pop, and Mihaela-Claudia Cercel. 2024.
\newblock \href {https://doi.org/10.18653/v1/2024.emnlp-main.933} {Investigating {Large} {Language} {Models} for {Complex} {Word} {Identification} in {Multilingual} and {Multidomain} {Setups}}.
\newblock In \emph{Proceedings of the 2024 {Conference} on {Empirical} {Methods} in {Natural} {Language} {Processing}}, pages 16764--16800, Miami, Florida, USA. Association for Computational Linguistics.

\bibitem[{Turpin et~al.(2023)Turpin, Michael, Perez, and Bowman}]{NEURIPS2023_ed3fea90}
Miles Turpin, Julian Michael, Ethan Perez, and Samuel Bowman. 2023.
\newblock \href {https://proceedings.neurips.cc/paper_files/paper/2023/file/ed3fea9033a80fea1376299fa7863f4a-Paper-Conference.pdf} {Language models don\textquotesingle t always say what they think: Unfaithful explanations in chain-of-thought prompting}.
\newblock In \emph{Advances in Neural Information Processing Systems}, volume~36, pages 74952--74965. Curran Associates, Inc.

\bibitem[{Wei et~al.(2022)Wei, Wang, Schuurmans, Bosma, Ichter, Xia, Chi, Le, and Zhou}]{wei_chain--thought_2022}
Jason Wei, Xuezhi Wang, Dale Schuurmans, Maarten Bosma, Brian Ichter, Fei Xia, Ed~H. Chi, Quoc~V. Le, and Denny Zhou. 2022.
\newblock Chain-of-thought prompting elicits reasoning in large language models.
\newblock In \emph{Proceedings of the 36th {International} {Conference} on {Neural} {Information} {Processing} {Systems}}, {NIPS} '22, pages 24824--24837, Red Hook, NY, USA. Curran Associates Inc.

\end{thebibliography}

\appendix

\clearpage
\section{Prompts}
\label{sec:prompts}

For the FLUTE dataset, we use the following prompt: \\

\noindent \texttt{``You will be given two sentences, a premise and a hypothesis. Respond with either `entailment' if the hypothesis follows from the premise, or `contradiction' otherwise.''} \\

\noindent For the other datasets, we use: \\

\noindent \texttt{``Predict whether the MWE given in the sentence being used idiomatically or literally. Respond `idiomatic' or `literal' respectively.''} \\

\noindent For using GPT-4o to extract the label, we use: \\

\noindent \texttt{``Given the following model output, predict what label the model was trying to assign out of \{option\_string\}. Only output either the word \{option\_string\}''} \\

\noindent When adding the definition as part of the prompt, we append:\\

\noindent \texttt{``Definition: \{definition\}''} \\

\section{Error Analysis Labelling}
\label{sec:error-analysis-labels}

\begin{table}[!h]\centering
\resizebox{\columnwidth}{!}{
\begin{tabular}{llr}
\toprule
Type & Label & Score \\
\midrule
\textbf{Understanding} & No definition  & 0 \\
& Focus on single word  & 0 \\
& Inaccurate definition  & 0 \\
& Partial definition  & 2 \\
& Accurate definition  & 3 \\
\midrule
\textbf{Reasoning} & Nonsensical reasoning &  0 \\
& OK reasoning &  2 \\
& Good reasoning &  3 \\
& Meta-reasoning (inaccurate) &  2 \\
& Meta-reasoning (accurate) &  3 \\
\bottomrule
\end{tabular}
}
\caption{The labels used for the manual labelling of the models CoT, with the scores assigned to each label when calculating averages.}\label{tab:accuracy_idiomatic_results}
\end{table}

\section{Limitations}
Whilst we do on average see a small improvement in performance for idiomaticity detection in the larger models, this result isn't wholly consistent, as the reasoning variant of Llama-3.3 70B performs worse by 0.037 F1 on the English subset of SemEval compared to the non-reasoning variant. The average performance increasing from reasoning is also lower for this model than Qwen2.5 14B and Qwen2.5 32B (0.013 compared to 0.039). Further work could investigate a reason for this inconsistency.

On our human analysis of the model CoTs, we break the evaluation into ``understanding'', which we take as the ability of a model to define the relevant PIE in its idiomatic sense, and ``reasoning'', which we take as the ability of a model to use the context and the PIE to determine whether it's literal or idiomatic. This provides a useful analysis, but a more fine-grained evaluation using other features would perhaps be more useful, for example looking at whether a model is logically consistent, whether it's hallucinating, etc.

\end{document}